# Circle detection using Discrete Differential Evolution Optimization

**[*]Erik Cuevas, [*]Daniel Zaldivar[1], [*]Marco Pérez-Cisneros and [+]Marte Ramírez-Ortegón**

[*]Departamento de Ciencias Computacionales
Universidad de Guadalajara, CUCEI
Av. Revolución 1500, Guadalajara, Jal, México
{erik.cuevas, [1]daniel.zaldivar, marco.perez}@cucei.udg.mx
[+]Freie Universität Berlin
Takustrs 9, PLZ 14195, Berlin, Germany

**Abstract**

This paper introduces a circle detection method based on Differential Evolution (DE) optimization. Just as circle detection has been lately considered as a fundamental component for many computer vision algorithms, DE has evolved as a successful heuristic method for solving complex optimization problems, still keeping a simple structure and an easy implementation. It has also shown advantageous convergence properties and remarkable robustness. The detection process is considered similar to a combinational optimization problem. The algorithm uses the combination of three edge points as parameters to determine circles candidates in the scene yielding a reduction of the search space. The objective function determines if some circle candidates are actually present in the image. This paper focuses particularly on one DE-based algorithm known as the Discrete Differential Evolution (DDE), which eventually has shown better results than the original DE in particular for solving combinatorial problems. In the DDE, suitable conversion routines are incorporated into the DE, aiming to operate from integer values to real values and then getting integer values back, following the crossover operation. The final algorithm is a fast circle detector that locates circles with sub-pixel accuracy even considering complicated conditions and noisy images. Experimental results on several synthetic and natural images with varying range of complexity validate the efficiency of the proposed technique considering accuracy, speed, and robustness.

## 1. Introduction

The problem of detecting circular features holds paramount importance for image analysis, in particular for industrial applications such as automatic inspection of manufactured products and components, aided vectorization of drawings, target detection and so [1]. Solving object location is commonly approached by two techniques: deterministic techniques which include the application of Hough-transform-based methods [2], geometric hashing and template or model matching techniques [3, 4]. On the other hand, stochastic techniques include random sample consensus techniques [5], simulated annealing [6] and genetic algorithms (GA) [7].

Template and model matching techniques have been successfully applied to shape detection among several other [8]. Shape coding techniques and combination of shape properties have also been used to represent such objects. The main drawback is related to the contour extraction step from real images as it is difficult, for some models, to deal with pose invariance except for very few simple objects.

Commonly, the circle detection in digital images is performed through the Circular Hough Transform [9]. A typical Hough-based approach employs an edge detector and uses edge information to infer locations and radius values. Peak detection is then performed by averaging, filtering and histogramming the transform space. However, such approach requires a large storage space given the 3-D cells needed to cover the parameters ($x$, $y$, $r$), the computational complexity and the low processing speed. Moreover, the accuracy of the extracted parameters of the detected circle is poor, particularly in presence of noise [10]. In particular for a digital image of significant width and height and densely populated edge pixels, the required processing time for Circular Hough Transform makes it prohibitive to be deployed in real time applications. In order to overcome such a problem, other researchers have proposed new methods whose principles differs from classic Hough transform although the name is somehow kept. for instance the

[1] Corresponding author, Tel +52 33 1378 5900, ext. 7715, E-mail: daniel.zaldivar@cucei.udg.mx

Probabilistic Hough Transforms [11], the Randomized Hough Transform (RHT) [12] and the Fuzzy Hough Transform [13]. Some other methods that incorporate alternative transforms have been reviewed by Becker at [14].

Stochastic search methods for the shape recognition in computer vision such as Genetic Algorithms (GA) have also offered important results. In particular, GA has recently been applied to shape detection through primitive extraction in the work by Roth and Levine in [15]. Lutton et al. have developed further improvement on the aforementioned method in [16]. Yao *et al.,* proposed a multi-population GA to detect ellipses [17]. In [49], GA has been used for template matching by applying an unknown affine transformation. Ayala–Ramirez *et al.,* presented a circle detector based on GA [18] which can detect multiple circles on real images but fails frequently on detecting imperfect circles. Recently, Das et al. in [19], have proposed an automatic circle detector (AnDE) through a novel optimization method which combines differential evolution and simulated annealing. The algorithm has been able to detect only one circle upon synthetic images. However, as the algorithm is based on a non-combinatorial approach (just like the original DE), it frequently converges to sub-optimal solutions. Moreover, unstable behaviors within the process of adapting a non-combinatorial approach in order to solve a combinatorial problem quite often yield a heavy computing load.

Another example is presented by Rosin et al. in [20]. It applies soft computing techniques to shape classification. In the case of ellipsoidal detection, Rosin proposes in [21] an ellipse fitting algorithm that uses five points. In [22], Zhang and Rosin extend the algorithm to fit data upon super-ellipses.

The novel evolutionary computation technique known as Differential Evolution (DE) has been introduced by Storn and Price in 1995 [23]. It has gained much attention yielding a wide range of applications for solving complex optimization problems. The procedure resembles the structure of an evolutionary algorithm (EA), but differs on the way it generates new candidate solutions and on the use of a ‘greedy’ selection scheme. Moreover, DE performs searching by using floating point representations opposing the binary wording, commonly used by other EA schemes. DE also takes relevant concepts for ‘larger populations’ from GAs, and ‘self-adapting mutation’ from ESs.

In general, the DE algorithm can be considered as a fast robust algorithm which represents an actual alternative to EA. Considering the advantages of DE over other optimization methods, it has become more and more popular in solving complex, nonlinear, non-differentiable and non-convex optimization problems. Over the recent years, DE has been successfully applied to different subjects such as reservoir system optimization [24], optimal design of shell-and-tube heat exchangers [25], beef property model optimization problems [26], generation planning problems [27], distribution network reconfiguration problems [28], capacitor placement problems [29], induction motor identification problems [30], optimal design of gas transmission network [31] and chaotic systems control and synchronization [32], just to name a few.

An optimization problem is considered to be combinatorial as long as its set of feasible solutions is both finite and discrete, i.e., enumerable; therefore the classical DE cannot be applied to combinatorial or permutative problems unless a modification is added [33]. The situation arises from the fact that crossover and mutation mechanisms invariably change any given value to a real number, leading to unstable behaviors [34] and sub-optimal solutions [35]. By truncating real-valued parameters to their nearest feasible value, DE has been applied to combinatorial tasks [19]. However, such solution itself has led to infeasible solutions and long compute times [33,34]. Over the years, some researchers have been working on DE combinatorial optimization [36, 37], concluding that DE may be appropriate for combinatorial optimization, as it seems effective and competitive in comparison to other related approaches. Some combinatorial-based DE optimization schemes have proved their effectiveness: the Discrete Differential Evolution (DDE) [38,44], the Relative Position Indexing Approach [39], the Smallest Position Value Approach [40], the Discrete/Binary Approach [41] and the Discrete Set Handling Approach [42]. Discrete Differential Evolution (DDE) has evolved from DE, offering better performance for solving combinatorial problems. In DDE, suitable conversion routines are incorporated into DE, aiming to operate from integer values to real values in such a way that integer values can be retrieved after the crossover operation.

This paper presents an algorithm for the automatic detection of circular shapes from complicated and noisy images with no consideration of the conventional Hough transform principles. The detection process is considered to be similar to a combinatorial optimization problem. The DDE is then applied to search the entire edge-map for circular shapes. The algorithm uses the combination of three non-collinear edge points as candidate circles (*x*,*y*,*r*) in the edge image of the scene. A new objective function to measure the existence of a candidate circle on the edge map has bee proposed following the neighbourhood of a central pixel through the Midpoint Circle Algorithm (MCA) [43]. Guided by the values of this objective function, the set of encoded candidate circles are evolved using the DDE so that they can fit into actual circles within the edge map of the image. The approach generates a sub-pixel circle detector which can effectively identify circles in real images despite circular objects exhibiting a significant occluded portion. Experimental evidence shows the effectiveness of such method for detecting circles under various conditions. A comparison to one state-of-the-art GA-based method [18] and the Annealed Differential Evolution Algorithm (AnDE) [19] on different images has been included to demonstrate the superior performance of the proposed method.

This paper is organized as follows: Section 2 provides a brief outline of the classical DE. Section 3 briefly discusses the combinatorial optimization problem and explains the DDE approach. In Section 4, the Circle detection algorithm based on DDE is assembled. Experimental results for the proposed approach are presented in Section 5 while Section 6 discusses some relevant conclusions.

## 2. Differential evolution algorithm

In recent years, there has been a growing research interest in the application of evolutionary algorithms for several fields of science and engineering. The differential evolution (DE) algorithm ([23], [45], [46]) is a relatively novel optimization technique that is able to efficiently solve numerical-optimization problems. The algorithm has successfully been applied in many different problems while it has gained a wide acceptance and popularity because of its simplicity, robustness, and good convergence properties [47].

The DE algorithm is a simple and direct search algorithm based on population that aims for optimizing global multi-modal functions. Like Genetic Algorithms (GA), it employs the crossover and mutation operators and the selection mechanism. An important difference between the GA and DE algorithms is that the GA relies on the crossover operator to provide the exchange of information among the solutions, building the best solutions. On the other hand, the DE algorithm relies on the mutation operation as the main operator and also employs a non-uniform crossover that takes child vector parameters from one parent more frequently than from others. The non-uniform crossover operator efficiently shuffles information about successful combinations. This enables the search to focus on the most promising area of the solution space.

The DE algorithm also introduces a novel mutation operation which is not only simple, but also significantly effective. It is based on the differences of randomly sampled pairs of solutions in the population. The DE algorithm is also fast, easy to use, very easily adaptable for integer and discrete optimization and quite effective for nonlinear constraint optimization including penalty functions.

The version of DE algorithm used in this work is known as DE/best/l/exp or ‘‘DE1” ([23], [45]). Classic DE algorithms begin by initializing a population of $N_p$ and *D*-dimensional vectors considering parameter values that are randomly distributed between the pre-specified lower initial parameter bound $x_{j,\text{low}}$ and the upper initial parameter bound $x_{j,\text{high}}$ as follows:

$$\begin{gathered} x_{j,i,t} = x_{j,\text{low}} + \text{rand}(0,1) \cdot (x_{j,\text{high}} - x_{j,\text{low}}); \\ j = 1,2,\ldots,D; \quad i = 1,2,\ldots,N_p; \quad t = 0. \end{gathered} \tag{1}$$

The subscript *t* is the generation index, while *j* and *i* are the parameter and population indexes respectively. Hence, $x_{j,i,t}$ is the *j*th parameter of the *i*th population vector in generation *t*. In order to generate a trial solution, DE algorithm first mutates the best solution vector $\mathbf{x}_{best,t}$ from the current population by adding the scaled difference of two vectors from the current population.

$$\begin{gathered} \mathbf{v}_{i,t} = \mathbf{x}_{best,t} + F \cdot (\mathbf{x}_{r_1,t} - \mathbf{x}_{r_2,t}); \\ r_1, r_2 \in \{1,2,\ldots,N_p\} \end{gathered} \tag{2}$$

with $\mathbf{v}_{i,t}$ being the mutant vector. Vector indices $r_1$ and $r_2$ are randomly selected except that they are all distinct and have no relation to the population index *i* (i.e., $r_1 \neq r_2 \neq i$). The mutation scale factor *F* is a positive real number that is typically less than one. Figure 1 illustrates the vector-generation process defined by Equation (2).

The next step in the crossover operation is that one or more parameter values of the mutant vector $\mathbf{v}_{i,t}$ are exponentially crossed with those belonging to the *i*th population vector $\mathbf{x}_{i,t}$. The result is the trial vector $\mathbf{u}_{i,t}$ which is computed by considering element to element as follows:

$$u_{j,i,t} = \begin{cases} v_{j,i,t}, & \text{if rand}(0,1) \leq Cr \text{ or } j = j_{\text{rand}}, \\ x_{j,i,t}, & \text{otherwise.} \end{cases} \tag{3}$$

with $j_{\text{rand}} \in \{1, 2, \ldots, D\}$.

The crossover constant $(0.0 \leq Cr \leq 1.0)$ controls the fraction of parameters to which the mutant vector is contributing in final trial vector. In addition, the trial vector always inherits the mutant vector parameter according to the randomly chosen index $j_{\text{rand}}$, assuring that the trial vector differs by at least one parameter from the vector to which it is being compared ( $\mathbf{x}_{i,t}$ ).

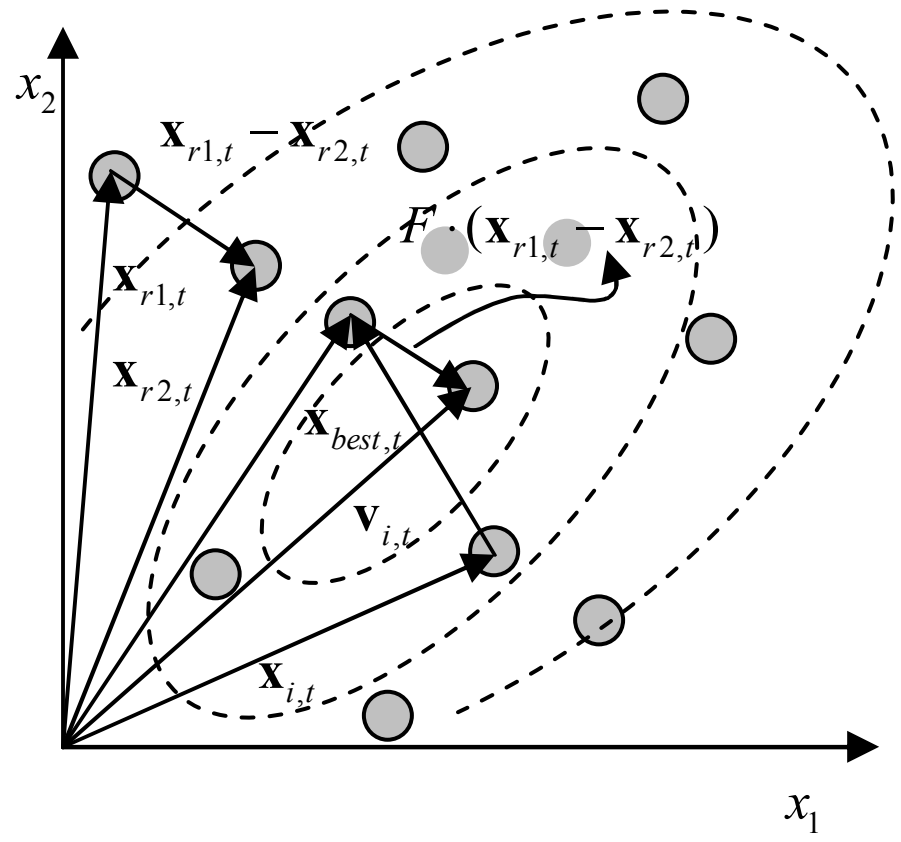

**Fig. 1.** Two-dimensional example of an objective function showing its contour lines and the process for generating **v** in scheme DE/best/l/exp from vectors of the current generation.

Finally, the selection operation is used to create the better solutions. Thus, if the computed cost function value for trial vector is less than or equal to target vector, then such trial vector replaces the target vector in the next generation. Otherwise, the target vector remains in the population for at least one more generation, yielding

$$\mathbf{x}_{i,t+1} = \begin{cases} \mathbf{u}_{i,t}, & \text{if } f(\mathbf{u}_{i,t}) \leq f(\mathbf{x}_{i,t}), \\ \mathbf{x}_{i,t}, & \text{otherwise.} \end{cases} \tag{4}$$

Here, *f* represents the cost function. These processes are repeated until a termination criterion is attained or a predetermined generation number is reached.

**3. Discrete Differential Evolution**

The classical DE cannot be applied directly to discrete, combinatorial or permutative problems unless it is modified. The crossover and mutation mechanisms invariably change any given value to a real number leading to unstable behaviors [34] and sub-optimal solutions [35]. By truncating real-valued parameters to their nearest feasible value, DE has been applied to combinatorial tasks [19]. However, such solution itself has led to unstable behaviors characterized by infeasible solutions and long compute times [33, 34].

Therefore it is required to modify the DE algorithm either changing the population mechanism or the crossover/mutation mechanism. In this work, the experiment followed the proposal given in [44]. It does not modify the DE strategies in any way, but manipulate the population in order to enable a smoother DE's operation. Since the solution for a given population is discrete in nature, a suitable conversion routine is required in order to change the solution from integer numbers to real-sized values, retrieving the integer–sized values after the crossover. The population is thus generated as a set of integers considering two conversions routines as follows: Forward transformation and backward transformation to convert between integer-size and real-size values accordingly. This new heuristic has been named as Discrete Differential Evolution (DDE) [38, 44].

The basic DDE outline is:

- ***Population Generation*****:** An initial discrete solutions number is generated for the initial population.
- ***Forward Transformation (Integer to Real Conversion)***: This conversion scheme transforms the solution from integer-size values to the real-size numbers.
- ***DE Strategy*****:** Full DDE algorithm described in section 3.
- ***Backward Transformation (Real to Integer Conversion)*****:** This conversion scheme transforms the solution back from real-size values to integer-sized numbers.
- ***Validation*****:** If the integer-size values represent a feasible solution, i.e. inside the interval of possible values, then it is accepted and evaluated by the next population.

*3.1 Forward Transformation.*

The conversion scheme transforms the solution from integer-size values to a real-size format. The formulation of the forward transformation is:

$$x_i' = -1 + \frac{x_i \cdot h \cdot 5}{10^3 - 1} \tag{5}$$

The equivalent real-size value $x_i'$ for $x_i$ is given as $10^2 \le 5 \cdot 10^2 < 10^3$. The domain of the variable $x_i$ is length = 5 as shown in $5 \cdot 10^2$. The precision of the value to be generated is set at two decimal places as it is given by the superscript two in $10^2$. The range of the variable $x_i$ falls between 1 and $10^3$. The lower bound is 1 whereas the upper bound of $10^3$ is obtained from extensive experimentation. The upper bound $10^3$ provides optimal filtering for closely generated values. The factor *h* has been established after extensive experimentation to 100 [44].

*3.2 Backward Transformation.*

The backward transformation is the reverse operation to forward transformation. It converts the real-size value back to an integer format as follows:

$$\text{int } [x_i'] = \frac{(1 + x_i') \cdot (10^3 - 1)}{5 \cdot h} \tag{6}$$

Assuming $x_i'$ to be the real-size value obtained from the classical DE algorithm. The value $x_i'$ is rounded to the nearest integer.

*3.3 Validation*

The DDE algorithm searches for an optimum within the feasible search-space *S,* just like any other stochastic optimization algorithm. However, in the scope of this paper, there exist some parameter sets that might not provide physically feasible solutions to the problem, despite they belong to the searching space. Moreover, restricting the search-space to the feasible region might be difficult because the constraints are not that simple to be established [46]. Therefore a penalty strategy [46, 47] should be implemented within the DE algorithm for tackling such a problem. If a candidate set is not a physically plausible solution, i.e. yielding an unstable system, then an exaggerated cost function value is returned. As this value is uncommonly large in comparison to usual cost function values, such ‘‘unstable” siblings are usually eliminated on a single generation.

## 4. Circle detection using DDE

Circles are represented in this work by means of parameters of the well-known second degree equation (see Equation 7), that passes through three points in the edge space of the image. Images are preprocessed by an edge detection step as a single-pixel edge detection method for object's contour is required. This task is accomplished by the classical Canny algorithm yielding the locations for each edge point. Such points are the only potential candidates to define circles by considering triplets. All the edge points in the image are thus stored in a vector array $P=\left\{p_1,p_2,\ldots,p_{N_p}\right\}$ with $N_p$ as the total number of edge pixels contained in the image. So the algorithm stores the $(x_i,y_i)$ coordinates for each edge pixel $p_i$ inside an edge vector.

In order to construct each of the circle candidates (or individuals within the DE framework), indexes $i_1$, $i_2$ and $i_3$ of three edge points must be combined, since the circle's contour is assumed to go through points $p_{i_1}$; $p_{i_2}$; $p_{i_3}$. A number of candidate solutions are generated randomly for the initial population. The solutions will thus evolve through the application of the DDE algorithm as the evolution over the population takes place until a minimum is reached and the best individual is considered as the solution for the circle detection problem.

Applying classic methods based on Hough Transform for circle detection would normally require huge amounts of memory and consume large computation time. In order to reach a sub-pixel resolution –an equal feature of the method presented in this paper, they also consider three edge points to cast a vote for the corresponding point within the parameter space. Such methods also require an evidence-collecting step that is also implemented by the method in this paper but as the evolution process is performed and the objective function improves at each generation by discriminating non plausible circles. Thus, the circle will be located, with no visits to several image points.

The following paragraphs clearly explain the required steps to formulate the circle detection task just as one DE optimization problem.

### *4.1. Individual representation*

Each element $C$ of the population uses three edge points as elements. In this representation, the edge points are stored according to one index that is relative to their position within the edge array $P$ of the image. That will encode an individual as the circle that passes through the three points $p_i$, $p_j$ and $p_k$ ($C=\{p_i,p_j,p_k\}$). Each circle $C$ is represented by three parameters $x_0$, $y_0$ and $r$, being $(x_0,y_0)$ the $(x, y)$ coordinates of the center of the circle and $r$ its radius. The equation of the circle passing through the three edge points can be computed as follows:

$$(x-x_0)^2+(y-y_0)^2=r^2 \tag{7}$$

considering

$$\mathbf{A}=\begin{bmatrix} x_j^2+y_j^2-(x_i^2+y_i^2) & 2\cdot(y_j-y_i) \\ x_k^2+y_k^2-(x_i^2+y_i^2) & 2\cdot(y_k-y_i)\end{bmatrix} \quad \mathbf{B}=\begin{bmatrix} 2\cdot(x_j-x_i) & x_j^2+y_j^2-(x_i^2+y_i^2) \\ 2\cdot(x_k-x_i) & x_k^2+y_k^2-(x_i^2+y_i^2)\end{bmatrix}, \tag{8}$$

$$x_0=\frac{\det(\mathbf{A})}{4((x_j-x_i)(y_k-y_i)-(x_k-x_i)(y_j-y_i))}, \; y_0=\frac{\det(\mathbf{B})}{4((x_j-x_i)(y_k-y_i)-(x_k-x_i)(y_j-y_i))}, \tag{9}$$

and

$$r = \sqrt{(x_0 - x_d)^2 + (y_0 - y_d)^2} \ , \tag{10}$$

with det(.) representing the determinant and $d \in \{i, j, k\}$. Figure 2 illustrates the parameters defined by Equations 7 to 10.

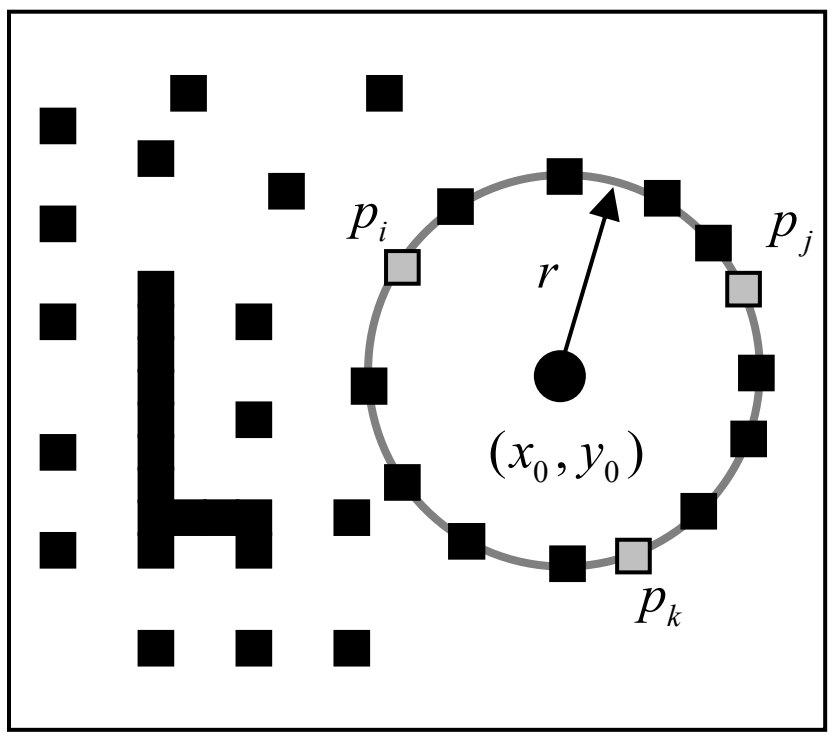

**Fig. 2.** Circle candidate (individual) built from the combination of points $p_i$, $p_j$ and $p_k$.

Thus, it is possible to represent the shape parameters (for the circle, [ $x_0$, $y_0$, $r$]) as a transformation $T$ of the edge vector indexes $i, j$ and $k$.

$$[x_0, y_0, r] = T(i, j, k) \tag{11}$$

with $T$ being the transformation calculated using $x_0$, $y_0$, and $r$ from equations (7) to (10).

By exploring each index as an individual parameter, it is possible to sweep the continuous space looking for the shape parameters using the DE optimization but keeping an integer representation through the DDE algorithm. This approach reduces the search space by eliminating unfeasible solutions.

*4.2 Objective function*

Optimization refers to the search of parameters that minimize an objective function or error. In order to calculate the error produced of an individual $C$, the circumference coordinate is calculated as a virtual shape. It must be thus validated, i.e. if it really exists in the edge image. The test for these points is $S = \{s_1, s_2, \ldots, s_{N_s}\}$, with $N_s$ representing the number of test points over which the existence of an edge point will be seek.

The test $S$ is generated by the midpoint circle algorithm [42]. The midpoint circle algorithm (MCA) is a method seeking the required points for drawing a circle. It requires as inputs only the radius $r$ and the center point $(x_0, y_0)$. The algorithm considers the circle equation $x^2 + y^2 = r^2$, with only the first octant. It draws a curve starting at point ($r$, 0) and proceeds upwards-left by using integer additions and subtractions. See full details in [48].

The MCA aims to calculate the required points $N_s$ in order to represent the circle considering coordinates $S = \{s_1, s_2, \ldots, s_{N_s}\}$. The algorithm is considered the quickest providing a sub-pixel precision [48]. However, in order to protect the MCA operation, it is important to assure that points lying outside the image plane must not be considered as they must be included in $N_s$ too.

The objective function $J(C)$ represents the error produced among the pixels $S$ for the circle candidate $C$, i.e. the pixels that actually exist in the edge image, yielding:

$$J(C) = 1 - \frac{\sum_{i=1}^{N_s} E(x_i, y_i)}{N_s} \tag{12}$$

where $E(x_i, y_i)$ is a function that verifies the pixel existence in $(x_i, y_i)$ and $Ns$ is the number of pixels lying in the circle's perimeter that correspond to *C,* currently under testing. The function $E(x_i, y_i)$ allows a small error between the position of the test edge pixel and the circumference point of $(x_i, y_i)$ calculated by the MCA operation. For the test, it is considered a *D* x *D* neighborhood *NB* with center at $(x_i, y_i)$. It serves as the test-region *NB* where the pixel existence is to be verified. Hence, $E(x_i, y_i)$ is defined as:

$$E(x_i, y_i) = \begin{cases} 1 & \text{if a pixel within } NB \text{ in } (x_i, y_i) \text{ is an edge point} \\ 0 & \text{otherwise} \end{cases} \tag{13}$$

Thus, Equation (12) accumulates the number of successful edge test-points (points in *S*) that are actually present in the edge image. After some trial and error, it has been determined that taking *D*=5 is generally enough for the circular detection task. Therefore the algorithm tries to minimize $J(C)$ since a smaller value implies a better response (minimum error) of the "circularity" operator. The optimization process can thus be stopped after either the maximum number of epochs is reached or a satisfactory error in $J(C)$ is found.

### *4.3. Implementation of DDE*

The implementation of DDE can be summarized into the following steps:

**Step 1:** Setting the DDE parameters. Initializing the population of *m* individuals where each decision variable $p_i$, $p_j$ and $p_k$ of $C_a$ is set randomly within the interval $[1, N_p]$. All values must be integers. Considering $a \in (1, 2, \ldots, m)$.

**Step 2:** Evaluating the objective value $J(C_a)$ for all *m* individuals, and determining the $\mathbf{x}_{best}$ showing the best objective value (minimum value).

**Step 3:** Converting the parameters from integer-size to real-size format using Equation (5).

**Step 4:** Performing mutation operation for each individual according to Equation (2), in order to obtain each individual's mutant.

**Step 5:** Performing crossover operation between each individual and its corresponding mutant, following Equation (3) to obtain each individual's trial.

**Step 6:** Converting the parameters from real-size to integer-size format using Equation (6).

**Step 7:** Evaluating the objective values (*J*(*C*)) of trial individuals.

**Step 8:** Performing selection between each individual and its corresponding trial value, following Equation (4), in order to generate new individuals for the next generation.

**Step 9:** Checking all individuals. If a candidate parameter set is not physically plausible, i.e. out of the range $[1, N_p]$, then an exaggerated cost function value is returned. This aims to eliminate ''unstable" individuals.

**Step 10:** Determining the best individual of the current new population by using the best objective value. If the objective value is better than the objective value of $C_{best}$, then $C_{best}$ must be updated following the objective value of the current best individual in Equation 2.

**Step 11:** If the stopping criterion is met, then the output $C_{best}$ is the solution (a circle contained in the image), otherwise go back to Step 3.

## 5. Experimental results

In order to evaluate the performance of the proposed circle detector, several experimental tests have been developed as follows:

(5.1) Circle detection
(5.2) Shape discrimination
(5.3) Multiple circle detection
(5.4) Circular approximation
(5.5) Approximation from occluded circles, imperfect circles or arc detection
(5.6) Performance comparison

Table 1 presents the parameters of the DDE algorithm for this work. Once determined experimentally, they are kept for all the test images through all experiments.

| $F$ | $Cr$ | Population size | The search space for each variable $p_i$ , $p_j$ and $p_k$ | $D$ |
|---|---|---|---|---|
| 0.25 | 0.80 | 30 | $[1, N_p]$ | 5 |

**Table 1.** DDE detector parameters

All the experiments are performed on a Pentium IV 2.5 GHz computer under C language programming. All the images are preprocessed by the standard Canny edge-detector using the image-processing toolbox for MATLAB R2008a.

*5.1 Circle localization*

5.1.1. Synthetic images

The experimental setup includes the use of 20 synthetic images of 200x200 pixels. Each image has been generated drawing only one circle, randomly located. Some of these images were contaminated adding noise to increase the complexity in the detection process. The parameters to be detected are the center of the circle position ($x$, $y$) and its radius ($r$). The algorithm is set to 100 epochs for each test image. In all the cases the algorithm is able to detect the parameters of the circle, even in presence of noise. The detection is robust to translation and scale conserving a reasonably low elapsed time (typically under 1ms). Figure 3 shows the results of the circle detection for two different synthetic images.

5.1.2. Natural images

This experiment tests the circle detection upon real-life images. Twenty five images of 640x480 pixels are used on the test. All are captured using digital camera under 8-bit color format. Each natural scene includes a circle shape among other objects. All images are preprocessed using an edge detection algorithm and then fed into the DE-based detector. Figure 4 shows a particular case from the 25 test images.

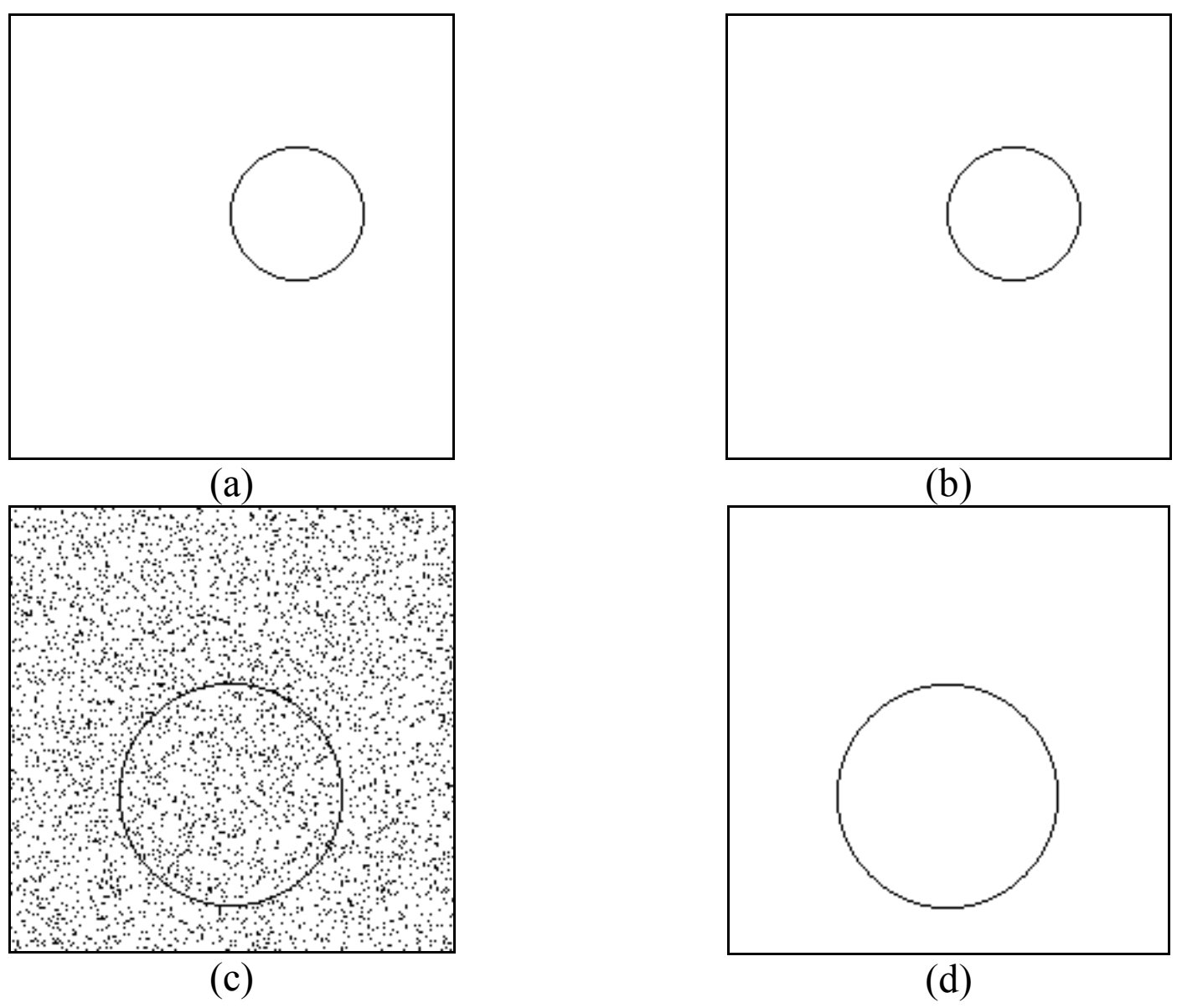

**Fig. 3.** Circle detection from synthetic images: (a) an original circle image, (b) its corresponding detected circle. (c) A second circle image with added noise and (d) its corresponding detected circle.

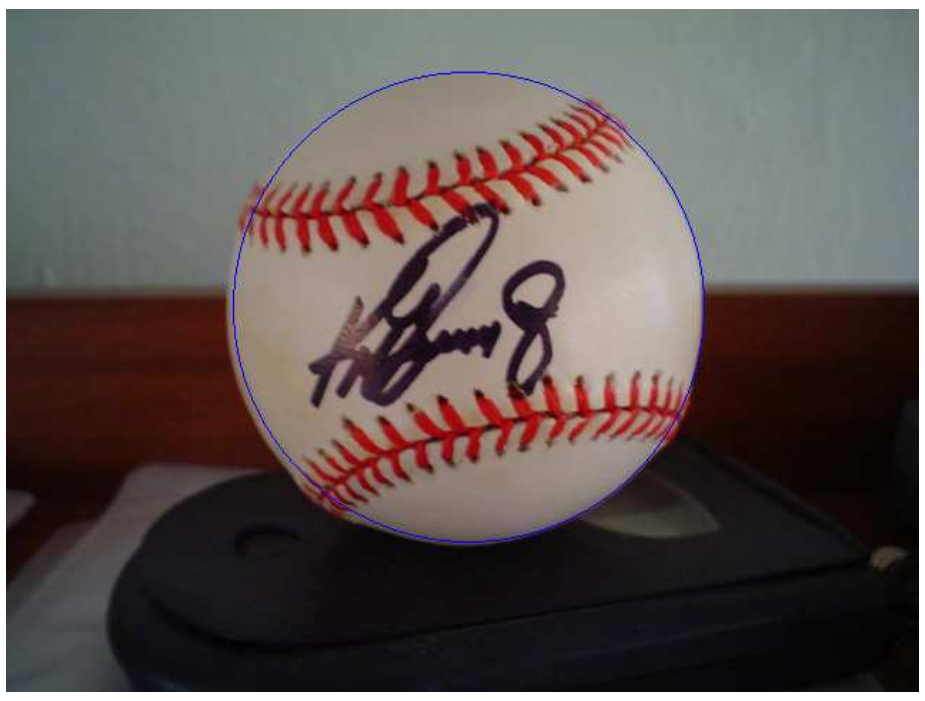

**Fig. 4.** Circle detection applied to a real-life image. The detected circle is shown near the ball's periphery.

Real-life images rarely contain perfect circles, and therefore the detection algorithm approximates the circle that better adapts to the imperfect circle within the noisy image. Such circle corresponds to the smallest error obtained for the objective function *J(C)*. The results on detection have been statistically analyzed for comparison purposes. For instance, the detection algorithm is executed 100 times on the same image (Figure 4), yielding the same parameters $x_0 = 210$, $y_0 = 325$, and *r*=165. This indicates that the proposed DDE algorithm is able to converge to the minimum solution obtained from the objective function *J(C)*. Again, the experiment has considered 100 epochs.

*5.2. Shape discrimination tests*

This section discusses on the circle detection ability when any other shapes are present in the image. Five synthetic images of 540x300 pixels are considered for the experiment. Noise has been added to all images. At each test, the algorithm runs 100 times storing two data: the number of correct circle detections and the elapsed time. A limit of 500 generations is set. The exact number of shapes per image is shown by Table 1. Figure 5 shows the algorithm being applied to the image number 5 which includes 11 shapes.

| Image | Shapes in the image | Time (s) | Accuracy rate (%) |
|---|---|---|---|
| 1 | 2 | 0.9 | 100 |
| 2 | 3 | 1.2 | 100 |
| 3 | 4 | 1.4 | 100 |
| 4 | 8 | 1.8 | 100 |
| 5 | 11 | 2.0 | 100 |

**Table 1.** Five test synthetic images are used for the shape discrimination experiment, considering five distinct numbers of shapes within the image.

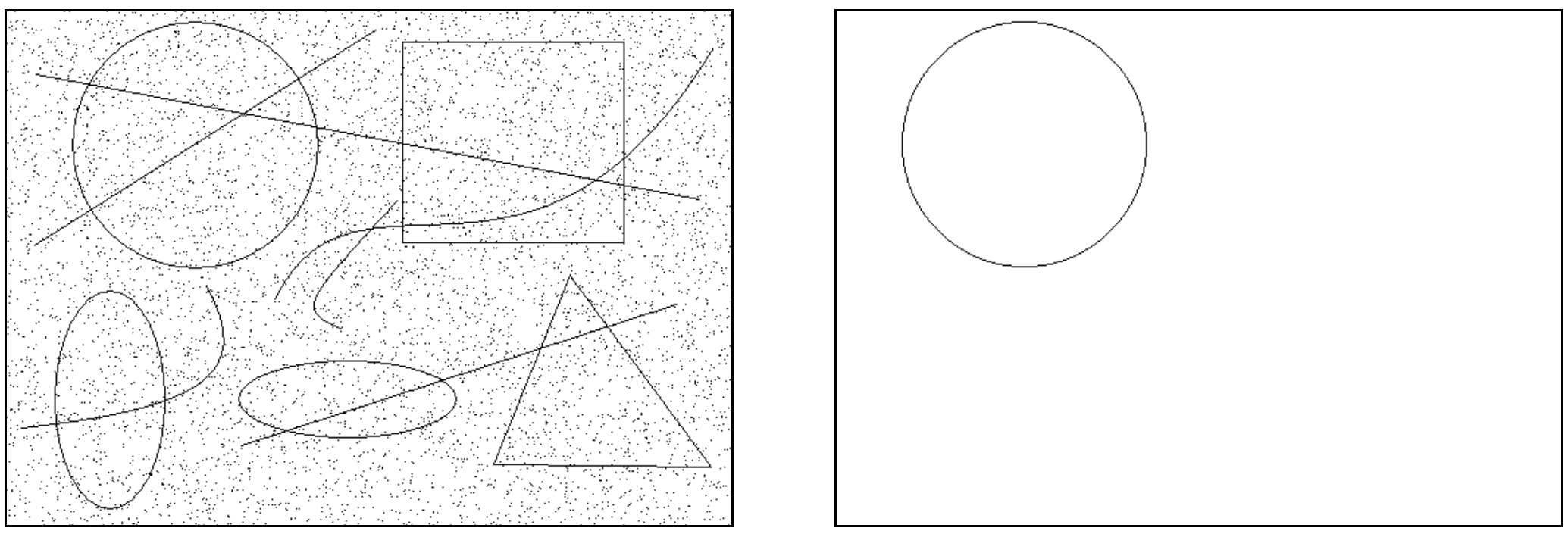

(a) (b)

**Fig. 5.** Results from testing eleven different shapes on a synthetic image: (a) the original image portraying added noise and (b) the detected circle

The same experiment is repeated using natural images. Table 2 resumes the system performance for this experiment. It is notorious that circles are fully detected for most cases and for about 97% in worst cases. That implies that the circle discrimination from other shapes is completely feasible on natural real-life images. Translation and change of scale are well handled by the approach. Figure 6 shows the algorithm being applied to the image number 4 which includes 5 different shapes.

| Image | Shapes in the image | Time (s) | Accuracy rate (%) |
|---|---|---|---|
| 1 | 2 | 0.9 | 100 |
| 2 | 3 | 0.9 | 100 |
| 3 | 4 | 0.9 | 100 |
| 4 | 5 | 1.8 | 100 |
| 5 | 8 | 3.2 | 97 |

**Table 2.** Shape Discrimination experiment using real-life images.

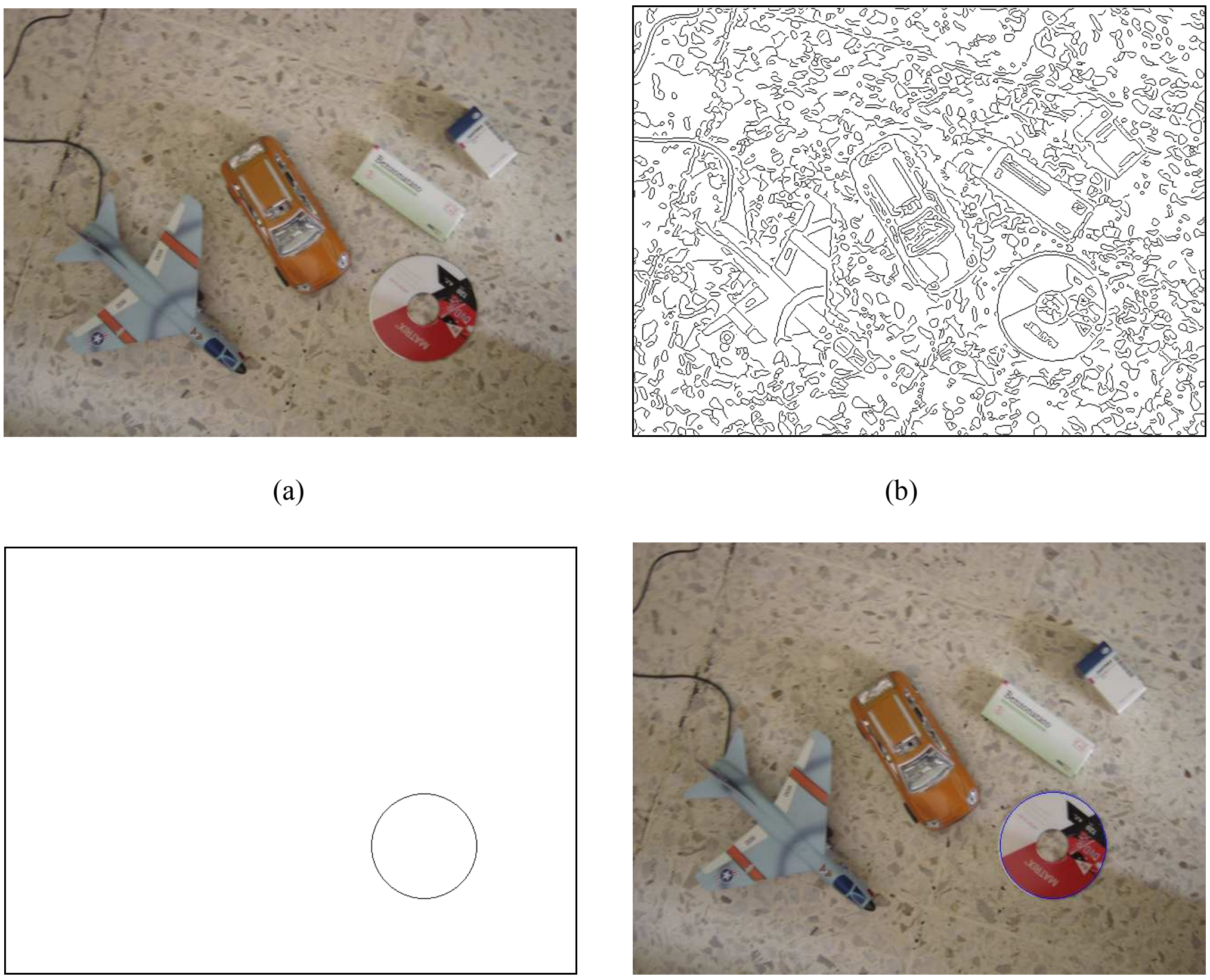

(a) (b)

(c) (d)

**Fig. 6.** Different shapes embedded into a real-life image. (a) The test image (b) the corresponding edge map, (c) circle detected and (d) the detected circle over the original image.

### *5.3. Multiple circle detection*

The approach is also capable of detecting several circles embedded into a real-life image. However the maximum number of circular shapes to be found must be set prior to operation. The approach will work over the edge image until the first circle is detected. Hence it represents the circle with the minimum objective function value $J(C)$. Once such shape is masked (i.e. eliminated) on the primary edge image, the DDE circle detector operates over the modified secondary image. The procedure is repeated until the maximum number of detected shapes is reached. Finally, a validation of all detected circles is performed by analyzing continuity of the detected circumference segments as proposed in [41]. Such procedure is required in case the user requests more circular shapes than the number of circles actually present in the image. If none of the detected shapes satisfies the circular completeness criterion, the system may simply reply a negative response such as: “no circle detected”. The algorithm may also identify any other circle-like shape in the image by selecting the best shapes until maximum shape number is reached.

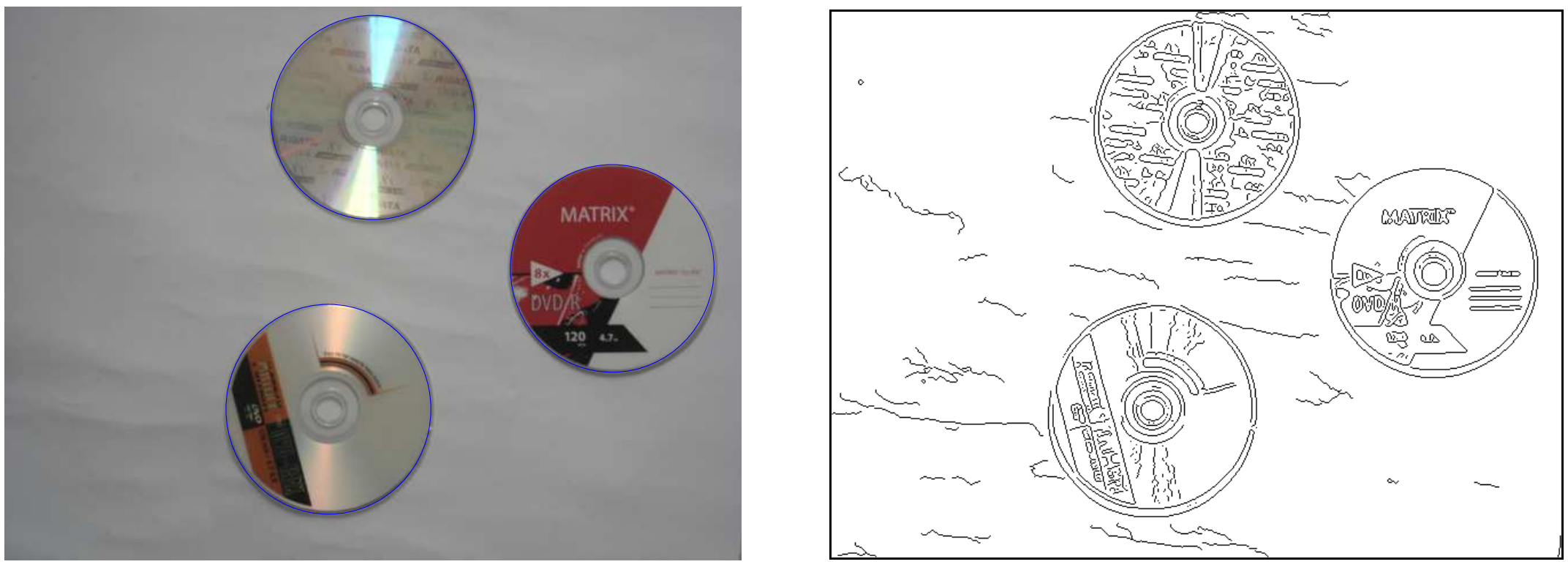

(a) (b)

**Fig. 7.** Multiple circle detection on real-life images: (a) the original image holding the best three detected circular shapes, (b) the edge image after applying the Canny algorithm.

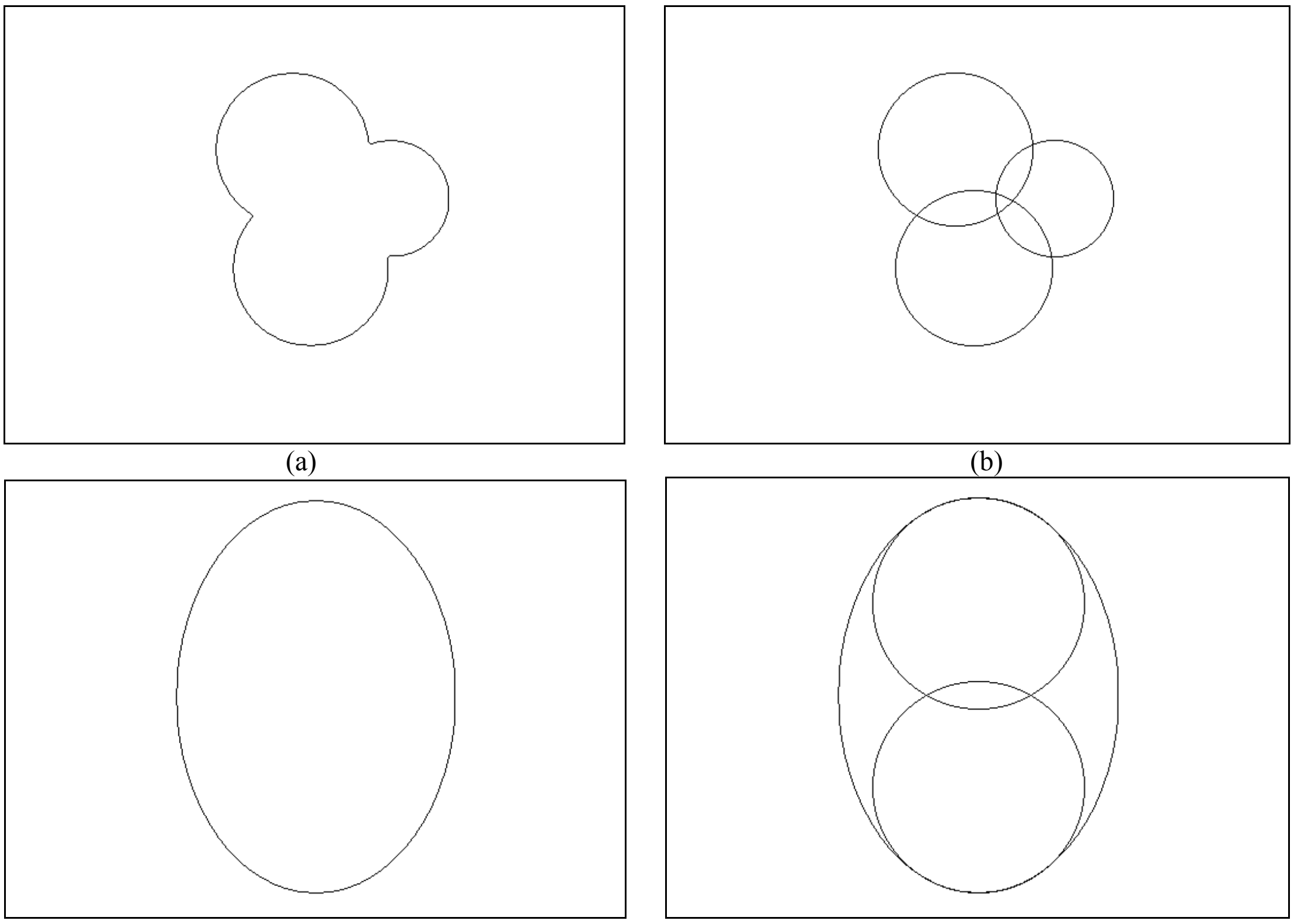

(a) (b) (c) (d)

**Fig. 8.** Circular approximation: (a) The original image (b) its circular approximation considering three circles, (c) original image and (d) its circular approximation considering only two circles.

Figure 7(a) shows a real-life image including several detected circles which have been previously sketched. For this case, the algorithm searched for the three best circular shapes. The edge image after the Canny algorithm application is also shown by Figure 7(b). Such image is the one actually fed to the algorithm.

The DDE algorithm is an iterative procedure which naturally builds an edge map containing the potential candidates that represent the circular shapes as they are identified through each step of the procedure. Each circle is rated according to the value recursively stored in the objective function $J(C)$, keeping track for the requested number of circles (such number being defined at the very beginning of the algorithm).

The iterative nature of the DDE algorithm means that the input image to the current step of the algorithm represents the optimized image from the previous step. The last image therefore does not include any fully detected circle because such a shape has already been registered into the edge map and no longer considered for future steps. The algorithm continues focusing exclusively on other potential maps which may or may not represent another circle. A maximum of 100 generations is normally considered as the limit for detecting a potential circle.

*5.4 Circular approximation*

Since circle detection has been considered an optimization problem, it is possible to approximate a given shape as the concatenation of circles. This can be achieved using one feature of the DDE algorithm which may detect multiple circles just as it was explained in the previous sub-section. Thus, the DDE algorithms may continually find circles which may approach a given shape according to the values already stored in the objective function *J(C).*

Figure 8 shows some examples of circular approximation. The applications of the circular approximation ranges from the detection of small circular segments of an object (see Figure 8(a) and 8(b)) up to the semi-detection of ellipses as it is shown by Figure 8(c) and 8(d).

*5.5.2 Circle extraction from occluded or imperfect circles and arc detection.*

The circle detection algorithm described in this paper may also be useful to approximate circular shapes from arc segments, occluded circular shapes or imperfect circles. This functionality is quite relevant considering such shapes are all common to typical computer vision problems. The proposed algorithm is able to find circle parameters that better approach to the arc, occluded or imperfect circles. Figure 9 shows some examples of this functionality. Recalling that, at this paper, the detection process is approached as an optimization problem and that the objective function *J(C)* gathers the *C* points actually contained on the image, it is evident that a smaller value of *J(C)* commonly refers to a circle while a greater value of *C* accounts either for an arc, an occluded circle or an imperfect circle. Such fact does not represent any trouble as circles would be detected first while other shapes would follow. In general, the detection of all kinds of circular shapes would only differ according to smaller or greater values of *J*(*C*).

*5.6. Performance comparison*

In order to enhance the algorithm analysis, the DDE algorithm is compared to the AnDE and the GA circle detectors over a set of common images.

For the GA algorithm described in Ayala-Ramirez et al.[18], the population size is 70, the crossover probability is 0.55, the mutation probability is 0.10 and number of elite individuals is 2. The roulette wheel selection and the 1-point crossover are both applied. The parameter setup and the fitness function follow the configuration suggested in [18]. The AnDE algorithm follows the implementation from [19], considering the experimental parameter set as: *F*=0.25, $Cr_{\max}=0.9$, $Cr_{\min}=0.7$, $\alpha=0.95$, $T_{\max}=100$.

Real-life images rarely contain perfectly-shaped circles. In order to test accuracy, the results are compared to a ground-truth circle which is manually detected from the original edge-map. The parameters $(x_{true}, y_{true}, r_{true})$ of such testing circle are computed using the Equations 7-10 for three circumference points from the manually detected circle. If the centre and the radius of such circle are successfully found by the algorithm by defining $(x_D, y_D)$ and $r_D$, then an error score can be defined as follows:

$$\text{Es}=\eta\cdot\left(|x_{true}-x_D|+|y_{true}-y_D|\right)+\mu\cdot|r_{true}-r_D| \tag{14}$$

The first term represents the shift of the centre of the detected circle as it is compared to the benchmark circle. The second term accounts for the difference between their radii. $\eta$ and $\mu$ are two weights associated to each term in Equation (14). They are chosen accordingly to agree the required accuracy as $\eta=0.05$ and $\mu=0.1$. This particular choice of parameters ensures that the radii difference would be strongly weighted in comparison to the difference of centre positions between the manually detected and the machine-detected circles.

In case the value **Es** is less than 1, the algorithm gets a success, otherwise it has failed in detecting the edge-circle. Notice that for $\eta = 0.05$ and $\mu = 0.1$, it yields **Es**<1 which accounts for a maximal tolerated difference on radius length of 10 pixels, whereas the maximum mismatch for the centre location can be up to 20 pixels. In general, the success rate (SR) can thus be defined as the percentage of reaching success after a certain number of trials.

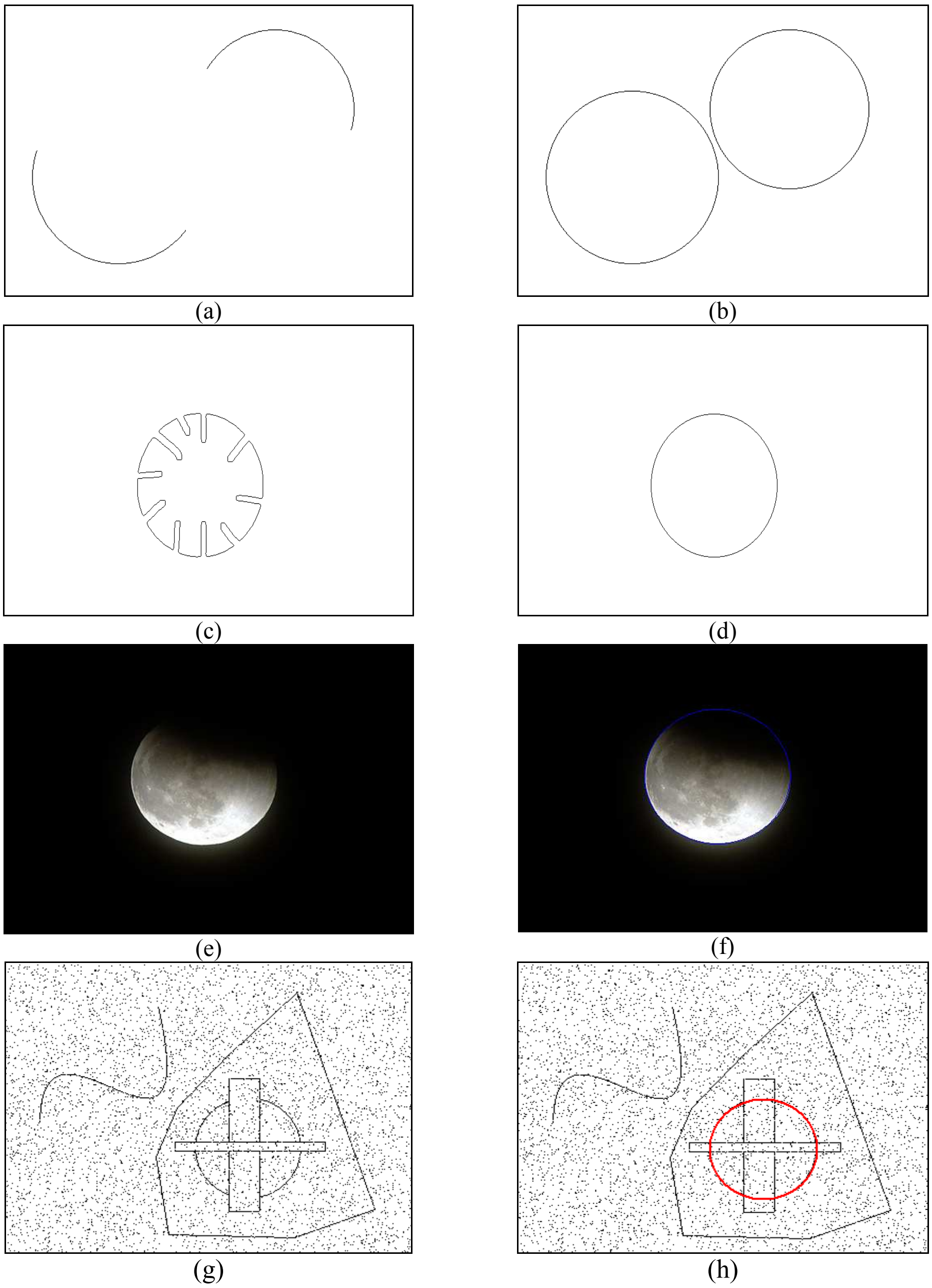

**Fig. 9.** Circle approximation from occluded shapes or imperfect circles and arc detection. (a) Original image portraying two arcs, (b) the circle approximation for the first image, (c) an imperfect circle, (d) the circle approximation for the imperfect circle, (e) an imperfect circle embedded into a natural image, (f) the circle approximation for the natural image, (g) an occluded circle and (h) its circle approximation.

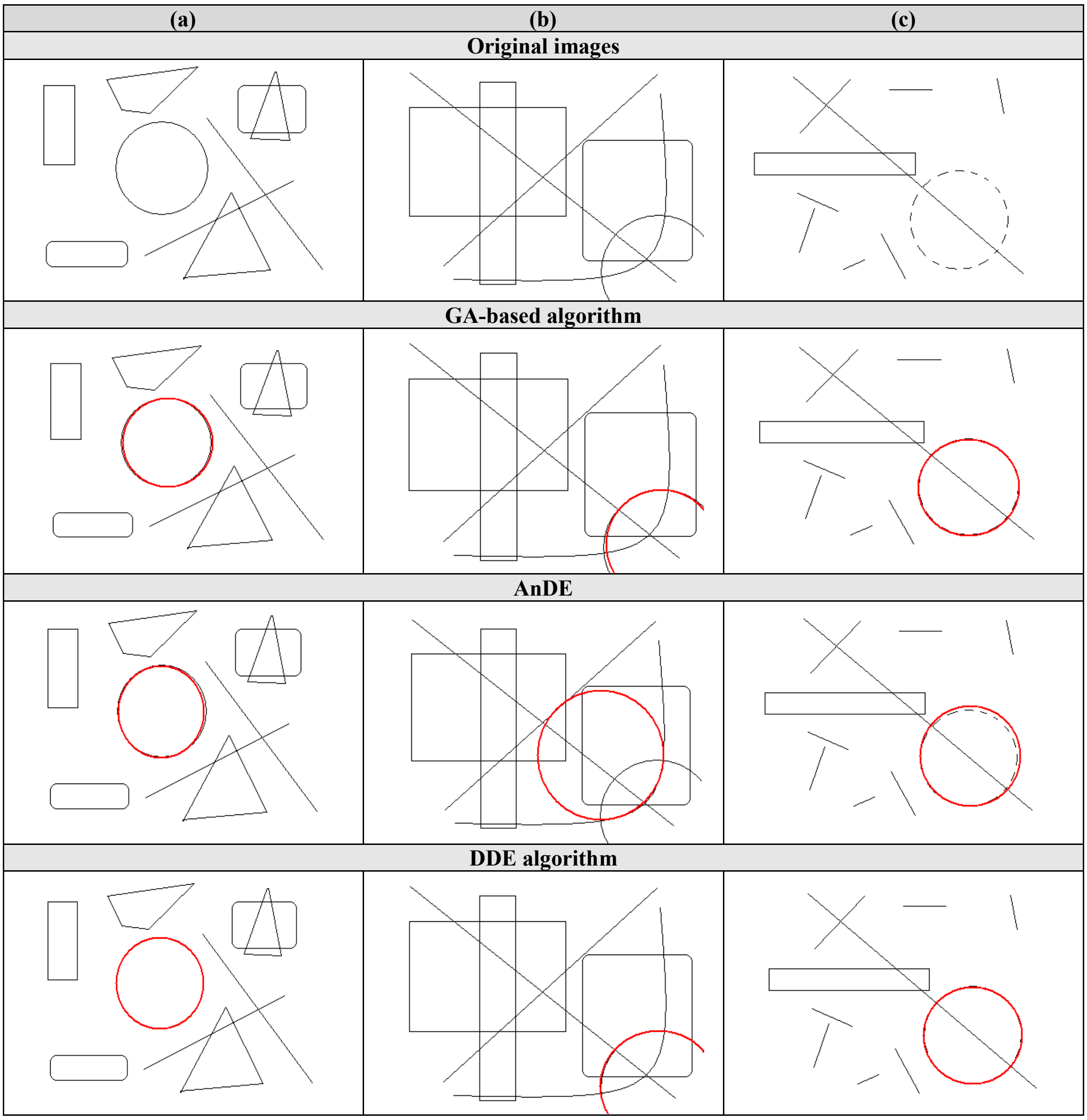

**Fig. 10.** Synthetic images and their corresponding detected circles.

| Image | Average time ± Standard deviation (s) | | | Success rate (SR) (%) | | | Es ± Standard deviation | | |
|---|---|---|---|---|---|---|---|---|---|
| | GA | AnDE | DDE | GA | AnDE | DDE | GA | AnDE | DDE |
| Synthetic images | | | | | | | | | |
| (a) | 1.37±(0.34) | 2.62±(0.92) | **0.30±(0.15)** | **100** | 90 | **100** | 0.61±(0.097) | 0.89±(0.091) | **0.31±(0.022)** |
| (b) | 1.61±(0.45) | 1.71±(0.87) | **0.34±(0.13)** | 90 | 83 | **100** | 0.68±(0.088) | 0.91±(0.088) | **0.27±(0.024)** |
| (c) | 2.71±(0.48) | 4.22±(0.82) | **0.41±(0.11)** | 88 | 80 | **99** | 0.76±(0.093) | 0.92±(0.086) | **0.19±(0.021)** |
| Natural Images | | | | | | | | | |
| (a) | 2.44±(0.51) | 4.11±(0.71) | **1.03±(0.37)** | 90 | 80 | **100** | 0.71±(0.098) | 0.92±(0.043) | **0.33±(0.039)** |
| (b) | 3.52±(0.60) | 6.78±(0.67) | **1.11±(0.29)** | 92 | 78 | **100** | 0.80±(0.095) | 1.02±(0.188) | **0.23±(0.031)** |
| (c) | 4.11±(0.58) | 7.88±(0.73) | **1.72±(0.31)** | 88 | 70 | **98** | 0.72±(0.111) | 1.42±(0.191) | **0.59±(0.044)** |

**Table 2.** The averaged execution-time and success rate of the GA-based algorithm, the AnDE method and the proposed DDE algorithm, considering six test images shown by Figures 10 and 11.

Figure 10 shows three synthetic images and the results obtained by the GA-based algorithm [18], the AnDE [19] and the proposed approach. Figure 11 presents the same experimental results considering real-life images. The results are averaged over 35 independent runs for each algorithm. Table 2 shows the averaged execution time, the success rate in percentage, and the averaged error score (**Es**) following Equation (22) for all three algorithms over six test images shown by Figures 10 and 11. The best entries are bold-cased in Table 2. A close inspection reveals that the proposed method is able to achieve the highest success rate and the smallest error, still requiring less computational time for most cases.

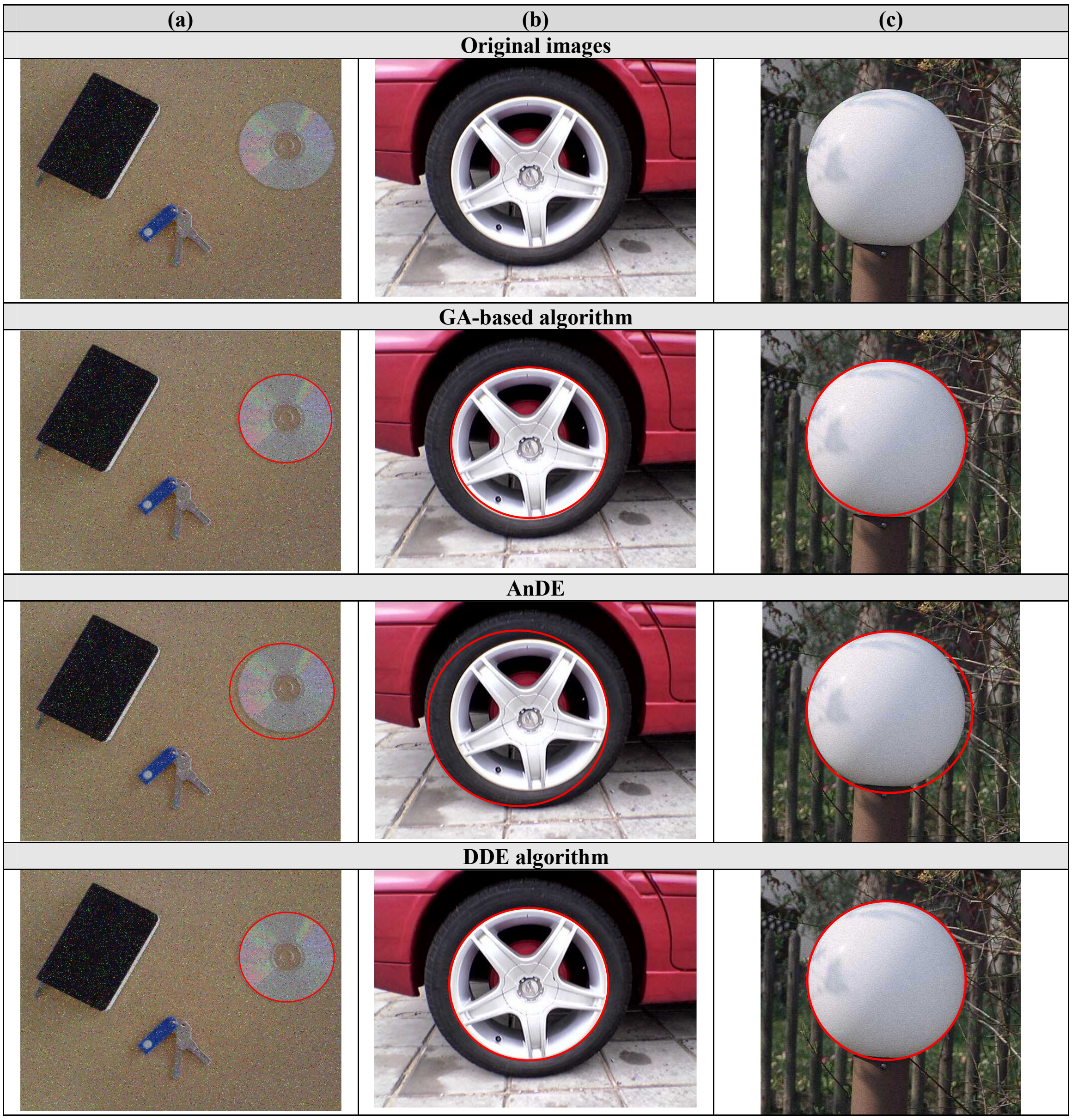

**Fig. 11.** Natural images and their detected circles used by the algorithm comparison.

## 6. Conclusions

This work has presented an algorithm for the automatic detection of circular shapes from complicated and noisy images with no consideration of the conventional Hough transform principles. The proposed method is based on the Discrete Differential Evolution algorithm (DDE) which in turn has demonstrated relevant improvements in comparison to the original Differential Evolution (DE) for solving combinatorial problems. The algorithm uses the combination of three non-collinear edge points as circle candidates within the edge image of the scene. An objective function evaluates if a given circle candidate is actually present in such edge image. Guided by the values of the objective function, the set of candidate circles are evolved using the DDE algorithm so that they can fit into the actual circles on the edge map of the image. As it can be verified from the results shown by Figures 3 to 11, our approach detects the circle in complex images with little visual distortion despite the presence of noisy background pixels.

Although the Hough Transform methods for circle detection also use three edge points to cast one vote for the potential circular point in the parameter space, they would require huge amounts of memory and longer computational time to obtain a sub-pixel resolution. Such methods also require an evidence-collecting step that has also been ultimately implemented by our method in a different pace as the evolution process is performed and the objective function improves at each generation by discriminating non plausible circles. Therefore a circle may be located avoiding visiting several image points.

An important contribution is the consideration of the overall circle detection process as a combinatorial optimization problem. Such view enables the algorithm to detect arcs or occluded circles still matching imperfect circles. The DDE is capable of finding circle parameters according to $J(C)$ instead of making a full review of all circle candidates towards detecting occluded or imperfect circles as it is commonly done by other methods.

A new objective function has been proposed in order to measure the existence of a candidate circle over the edge map of the image. It is based on the neighbourhood of a central pixel by means of the Midpoint Circle Algorithm (MCA) [39]. It is important to notice that at this paper, the use of the MCA exhibits more precision and confidence than those presented in [18] and [19] which only employ a set of candidate points extracted from the sampling of the circular pattern with some chances of missing an actual circle.

In order to test the circle detection accuracy, a score function is defined by Equation (14). It can effectively evaluate the mismatch between a manually detected circle and a machine-detected shape. We demonstrated that the DDE method outperforms both the GA (as described in [18]) and the AnDE (as described in [19]) within a statistically significant framework.

Although Table 2 indicates that the DDE method can yield better results on complicated and noisy images in comparison to the GA (as described in [18]) and the AnDE methods, notice that the aim of our paper is not intended to beat all the circle detector methods proposed earlier, but to show that the DDE algorithm can effectively serve as an attractive method to successfully extract circular shapes in images.

**References**

[1] da Fontoura Costa, L., Marcondes Cesar Jr., R. Shape Análisis and Classification. CRC Press, Boca Raton FL. (2001).

[2] Yuen, H., Princen, J., Illingworth, J., Kittler, J. Comparative study of Hough transform methods for circle finding. Image Vision Comput. 8 (1), 71–77. (1990).

[3] Iivarinen, J., Peura, M., Sarela, J., Visa, A. Comparison of combined shape descriptors for irregular objects. In: Proc. 8th British Machine Vision Conf., Cochester, UK, pp. 430–439. (1997).

[4] Jones, G., Princen, J., Illingworth, J., Kittler, J. Robust estimation of shape parameters. In: Proc. British Machine Vision Conf., pp. 43– 48. (1990).

[5] Fischer, M., Bolles, R. Random sample consensus: A paradigm to model fitting with applications to image analysis and automated cartography. CACM 24 (6), 381–395. (1981).

[6] Bongiovanni, G., and Crescenzi, P.: Parallel Simulated Annealing for Shape Detection, Computer Vision and Image Understanding, vol. 61, no. 1, pp. 60-69. (1995).

[7] Roth, G., Levine, M.D. Geometric primitive extraction using a genetic algorithm. IEEE Trans. Pattern Anal. Machine Intell. 16 (9), 901–905. (1994).

[8] Peura, M., Iivarinen, J., 1997. Efficiency of simple shape descriptors. In: Arcelli, C., Cordella, L.P., di Baja, G.S. (Eds.), Advances in Visual Form Analysis. World Scientific, Singapore, pp. 443–451.

[9] Muammar, H., Nixon, M. Approaches to extending the Hough transform. In: Proc. Int. Conf. on Acoustics, Speech and Signal Processing ICASSP_89, vol. 3, pp. 1556–1559. (1989).

[10] Atherton, T.J., Kerbyson, D.J. Using phase to represent radius in the coherent circle Hough transform, Proc, IEE Colloquium on the Hough Transform, IEE, London. (1993).

[11] Shaked, D., Yaron, O., Kiryati, N. Deriving stopping rules for the probabilistic Hough transform by sequential analysis. Comput. Vision Image Understanding 63, 512–526. (1996).

[12] Xu, L., Oja, E., Kultanen, P. A new curve detection method: Randomized Hough transform (RHT). Pattern Recognition Lett. 11 (5), 331–338. (1990).

[13] Han, J.H., Koczy, L.T., Poston, T. Fuzzy Hough transform. In: Proc. 2nd Int. Conf. on Fuzzy Systems, vol. 2, pp. 803–808. (1993).

[14] Becker J., Grousson S., Coltuc D. From Hough transforms to integral transforms. In: Proc. Int. Geoscience and Remote Sensing Symp., 2002 IGARSS_02, vol. 3, pp. 1444–1446. (2002).

[15] Roth, G. and Levine, M. D.: Geometric primitive extraction using a genetic algorithm. *IEEE Trans.Pattern Anal. Machine Intell*. 16 (9), 901–905. (1994).

[16] Lutton, E., Martinez, P.: A genetic algorithm for the detection 2-D geometric primitives on images. In: *Proc. of the 12th Int. Conf. on Pattern Recognition*, vol. 1, pp. 526–528. (1994).

[17] Yao, J., Kharma, N., and Grogono, P.: Fast robust GA-based ellipse detection. *In: Proc. 17th Int. Conf. on Pattern Recognition ICPR-04*, vol. 2, Cambridge, UK, pp. 859–862. (2004).

[18] Ayala-Ramirez, V., Garcia-Capulin, C. H., Perez-Garcia, A., and Sanchez-Yanez, R. E.: Circle detection on images using genetic algorithms, *Pattern Recognition Letters*, 27, 652–657. (2006).

[19] Das Swagatam, Dasgupta Sambarta, Biswas Arijit, Abraham Ajith: Automatic Circle Detection on Images with Annealed Differential Evolution. 8th International Conference on Hybrid Intelligent Systems 2008: 684-689.

[20] Rosin, P.L., Nyongesa, H.O. Combining evolutionary, connectionist, and fuzzy classification algorithms for shape analysis. In: Cagnoni, S. et al. (Eds.), Proc. EvoIASP, Real-World Applications of Evolutionary Computing, pp. 87–96. (2000).

[21] Rosin, P.L. Further five point fit ellipse fitting. In: Proc. 8th British Machine Vision Conf., Cochester, UK, pp. 290–299. (1997).

[22] Zhang, X., Rosin, P.L. Superellipse fitting to partial data. Pattern Recognition 36, 743–752. (2003).

[23] Storn R, Price K. Differential evolution – a simple and efficient adaptive scheme for global optimization over continuous spaces. Technical Rep. No. TR-95-012, International Computer Science Institute, Berkley (CA). (1995).

[24] Reddy J.M, Kumar N.D. Multiobjective differential evolution with application to reservoir system optimization. J Comput Civil Eng, 21(2):136–46. (2007).

[25] Babu B, Munawar S. Differential evolution strategies for optimal design of shell-and-tube heat exchangers. Chem Eng Sci. 62(14):3720–39. (2007).

[26] Mayer D, Kinghorn B, Archer A. Differential evolution – an easy and efficient evolutionary algorithm for model optimization. Agr Syst, 83:315–28. (2005).

[27] Kannan S, Mary Raja Slochanal S, Padhy N. Application and comparison of metaheuristic techniques to generation expansion planning problem. IEEE Trans Power Syst, 20(1):466–75. (2003).

[28] Chiou J, Chang C, Su C. Variable scaling hybrid differential evolution for solving network reconfiguration of distribution systems. IEEE Trans Power Syst, 20(2):668–74. (2005).

[29] Chiou J, Chang C, Su C. Ant direct hybrid differential evolution for solving large capacitor placement problems. IEEE Trans Power Syst, 19(4):1794–800. (2004).

[30] Ursem R, Vadstrup P. Parameter identification of induction motors using differential evolution. In: Proceedings of the 2003 congress on evolutionary computation (CEC'03), vol. 2, Canberra, Australia,. p. 790–6. (2003).

[31] Babu B, Angira R, Chakole G, Syed Mubeen J. Optimal design of gas transmission network using differential evolution. In: Proceedings of the second international conference on computational intelligence, robotics, and autonomous systems (CIRAS-2003), Singapore. (2003).

[32] Zelinka, I., Chen, G., Celikovsky, S.: Chaos sythesis by means of evolutionary algorithms. Int. J. Bifurcat Chaos 4, 911–942 (2008).

[33] Onwubolu, G. and Davendra D. Differential Evolution: A Handbook for Global Permutation-Based Combinatorial Optimization. Springer-Verlag, Heidelberg (2009).

[34] Yuan X., Su A., Nie H., Yuan Y., Wang L. Application of enhanced discrete differential evolution approach to unit commitment problem. Energy Conversion and Management, 50(9), 2009, pp. 2449-2456.

[35] Wang L., Pan Q.-K, Suganthan P.N., Wang W.-H., Wang Y.-M. A novel hybrid discrete differential evolution algorithm for blocking flow shop scheduling problems. Computers & Operations Research, 37(3), 2010, pp. 509-520

[36] Tasgetiren M.F., Pan Q.-K., Liang Y.-C. A discrete differential evolution algorithm for the single machine total weighted tardiness problem with sequence dependent setup times. Computers & Operations Research, 36(6), 2009, pp. 1900-1915

[37] Tasgetiren M.F., Suganthan P.N., Pan Q.-K. An ensemble of discrete differential evolution algorithms for solving the generalized traveling salesman problem. Applied Mathematics and Computation, 215(9), 2010, pp. 3356-3368

[38] Onwubolu, G., Davendra, D.: Scheduling flow shops using differential evolution algorithm. Eur. J. Oper. Res. 171, 674–679 (2006).

[39] Lichtblau D.: Relative Position Index Approach. Davendra, D. and Onwubolu G (eds.) Differential Evolution A Handbook for Global Permutation-Based Combinatorial Optimization, pp. 81–120. Springer, Heidelberg (2009).

[40] Tasgetiren F., Chen A., Gencyilmaz G., Gattoufi S. Smallest Position Value Approach. Davendra, D. and Onwubolu G (eds.) Differential Evolution A Handbook for Global Permutation-Based Combinatorial Optimization, pp. 81–120. Springer, Heidelberg (2009)

[41] Tasgetiren F., Liang Y., Pan Q. and Suganthan P.: Discrete/Binary Approach. Davendra, D. and Onwubolu G (eds.) Differential Evolution A Handbook for Global Permutation-Based Combinatorial Optimization, pp. 81–120. Springer, Heidelberg (2009).

[42] Zelinka I., SDiscrete Set Handling. Davendra, D. and Onwubolu G (eds.) Differential Evolution A Handbook for Global Permutation-Based Combinatorial Optimization, pp. 81–120. Springer, Heidelberg (2009).

[43] Bresenham J.E. A Linear Algorithm for Incremental Digital Display of Circular Arcs. Communications of the ACM 20, 100-106. (1987).

[44] Davendra, D. and Onwubolu G.: Forward Backward Trasformation. Davendra, D. and Onwubolu G (eds.) Differential Evolution A Handbook for Global Permutation-Based Combinatorial Optimization, pp. 37–78. Springer, Heidelberg (2009).

[45] Reddy J.M, Kumar N.D. Multiobjective differential evolution with application to reservoir system optimization. J Comput Civil Eng, 21(2):136–46. (2007).

[46] Franco G, Betti R, Lus H. Identification of structural systems using an evolutionary strategy. Eng Mech, 130(10):1125–39. (2004).

[47] Koziel S, Michalewicz Z. Evolutionary algorithms, homomorphous mappings, and constrained parameter optimization. Evol Comput, 7(1):19–44. (1999).

[48] Van Aken, J.R., An Efficient Ellipse Drawing Algorithm, CG&A, 4(9), September 1984, pp 24-35. (1984).

[49] Yuen, S., Ma, C. Genetic algorithm with competitive image labelling and least square. Pattern Recognition 33, 1949–1966. (2000).